\documentclass[10pt,twocolumn,letterpaper]{article}
\usepackage[accsupp]{axessibility}
 \usepackage{iccv}              
\usepackage[most]{tcolorbox} 
\usepackage{amsthm}
\newtheorem{proposition}{Proposition}[section]

\usepackage{pifont}
\usepackage{times}
\usepackage[T1]{fontenc}
\usepackage{amssymb}

\usepackage{comment}
\usepackage{adjustbox}
\usepackage{enumitem}
\usepackage{amsmath}
\usepackage{algorithm}
\usepackage{varwidth}
\usepackage{algpseudocode}
\usepackage{graphicx}
\usepackage{float} 
\usepackage{stfloats}
\DeclareMathOperator*{\softmax}{softmax}
\usepackage{multirow}

\definecolor{iccvblue}{rgb}{0.21,0.49,0.74}
\usepackage[pagebackref,breaklinks,colorlinks,allcolors=iccvblue]{hyperref}
\usepackage[capitalize]{cleveref}
\crefname{figure}{Fig.}{Figs.}
\crefname{equation}{Eq.}{Eqs.}
\def\paperID{1138} 
\def\confName{ICCV}
\def\confYear{2025}

\title{Always Skip Attention}

\author{Yiping Ji$^{1,2}$, Hemanth Saratchandran$^{1}$, Peyman Moghadam$^{2,3}$, Simon Lucey$^1$\\
 $^1$Adelaide University,  $^2$CSIRO, $^3$Queensland University of Technology 
\\
{\tt\small \{yiping.ji, hemanth.saratchandran, simon.lucy\}@adelaide.edu.au,} \\
{\tt\small \{yiping.ji, peyman.moghadam\}@csiro.au}
}

\begin{document}
\twocolumn[{%
\renewcommand\twocolumn[1][]{#1}%
\maketitle
\begin{center}
    \centering
    \captionsetup{type=figure}
    \vspace{-0.7cm}
    \includegraphics[width=1\linewidth]{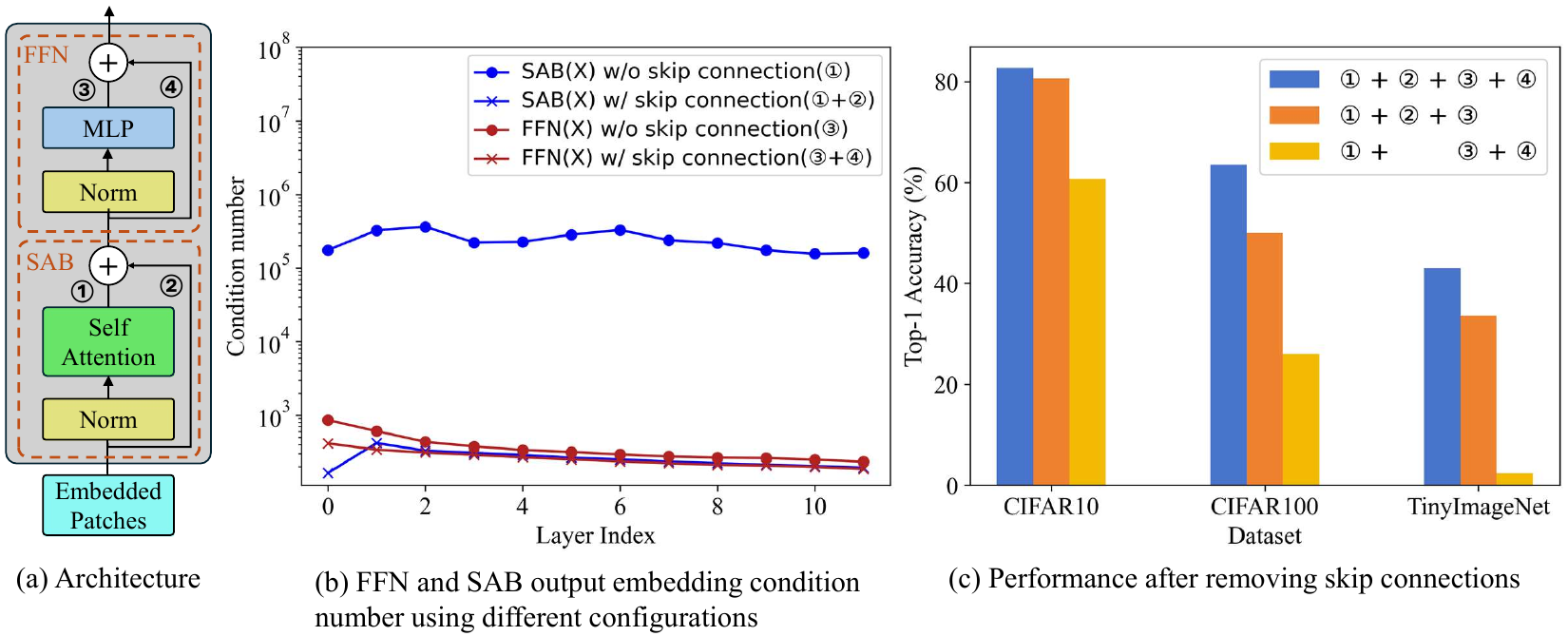}
\captionof{figure}{Removing skip connections from the Self-Attention Block (SAB) versus the Feedforward Network (FFN) in Vision Transformer (ViT) models results in different performance drop. \textbf{(a)} Model architecture of ViT consisting of a SAB, and a FFN in each layer. \ding{172} represents the output of the self-attention transformation, while \ding{174} represents the output of the MLP transformation. \ding{173} and \ding{175} represent identity mappings, referred to as skip connections, in the SAB and FFN, respectively.
\textbf{(b)} The condition number of the different output embeddings for each hidden layer. A ViT-Tiny model is trained in a typical manner with skip connections on both SAB and FFN blocks (\ie, \ding{172} + \ding{173} + \ding{174} + \ding{175}). It shows that the SAB output embedding without skip connection is poorly conditioned (\ding{172}), with a condition number significantly higher than that of the other three configurations involving ViT blocks.
\textbf{(c)} Classification results on three different datasets using the ViT-Tiny model under three different architectures at train time. The result shows the effects of removing skip connections at train time on the SAB and FFN blocks (either \ding{173} or \ding{175}) in all layers of ViT-Tiny.
    We observe that removing \ding{173} results in a catastrophic performance drop, whereas removing \ding{175} (as predicted) has a modest impact on performance.
This difference becomes more pronounced as the scale of the dataset increases.}
    \vspace{0.2cm}

    \label{fig:front_fig}
\end{center}%
}]

\maketitle

\begin{abstract}

We highlight a curious empirical result within modern Vision Transformers (ViTs). Specifically, self-attention catastrophically fails to train unless it is used in conjunction with a skip connection. This is in contrast to other elements of a ViT that continue to exhibit good performance (albeit suboptimal) when skip connections are removed. Further, we show that this critical dependence on skip connections is a relatively new phenomenon, with previous deep architectures (\eg, CNNs) exhibiting good performance in their absence. In this paper, we theoretically characterize that the self-attention mechanism is fundamentally ill-conditioned and is, therefore, uniquely dependent on skip connections for regularization. Additionally, we propose \textbf{T}oken \textbf{G}raying (\textbf{TG}), a simple yet effective complement (to skip connections) that further improves the condition of input tokens. We validate our approach in both supervised and self-supervised training methods.
\end{abstract}

\vspace{-0.5cm}
\section{Introduction}
Vision Transformers (ViTs) have demonstrated impressive performance on various computer vision and robotic tasks, such as object detection, semantic segmentation, video understanding, visual-language learning, and many others~\cite{oquab2023dinov2,zhou2024navgpt,liu2021swin,he2022masked, hausler2025pair}.
Much of this success is attributed to the self-attention mechanism, commonly referred to as Self-Attention Block (SAB), which allows ViTs to selectively attend to relevant parts of the input sequence when generating each element of the output sequence. 
Another essential component is the MLP, or Feedforward Networks (FFN), which facilitates intra-token communication across channel dimensions (see \cref{fig:front_fig}). Finally, identity mappings, commonly referred to as skip connections, are incorporated into both the SAB and FFN to enhance model performance.

We argue that the Jacobian of the SAB is disproportionately ill-conditioned compared to other components, notably the FFN block. 
A poorly conditioned~\footnote{The condition number of a full-rank matrix is defined by the ratio of the largest singular value to the smallest singular value. A lower condition number implies better condition.} Jacobian is fundamentally detrimental to gradient descent training
~\cite{pascanu2013difficultytrainingrecurrentneural, saratchandran2024weight,saratchandran2025enhancing,ji2025efficient, albert2025randlora}, impeding both convergence and stability. A natural strategy to address this issue is to explicitly model the Jacobian of the entire network. However, this approach is impractical due to the sheer scale of modern transformer architectures. To proceed, we introduce the following simplifying assumptions: 
\begin{tcolorbox}[grow to left by=0.01\linewidth, width=1\linewidth, boxsep=0mm, arc=2mm, left=2mm, right=2mm, top=2mm, bottom=2mm]
    
\textbf{Assumptions}

1. The condition of the network Jacobian is bound by the most ill-conditioned sub-block Jacobian.  

2. A proxy for the condition of the sub-block Jacobian is the condition of the output embedding of that block. 

\end{tcolorbox}

Although we do not provide formal proofs for these assumptions, we offer indirect validation through extensive empirical analysis.
First, we show that better conditioning of a sub-block’s output embeddings consistently correlates with improved empirical performance across multiple ViT benchmarks. 
Second, we demonstrate that this improvement in embedding condition is accompanied by a measurable enhancement in the sub-block's Jacobian condition 

\vspace{0.1cm}
\noindent \textbf{Skip in Transformers.}
Skip connections have become an indispensable component in modern transformer architectures~\cite{vaswani2017attention} and have been empirically proven to be a \textit{de facto} component.
In Fig.~\ref{fig:front_fig}~(b), we analyze a pre-trained ViT-Tiny model by measuring the condition numbers of the SAB and FFN output embeddings, both with and without skip connections.
%
Our observations reveal that the SAB output embedding becomes highly ill-conditioned in the absence of skip connections, with a condition number approaching $e^6$. 
In contrast, the other three configurations have relatively low condition numbers, around $e^3$. 
In other words, the SAB output embedding without skip connections has significantly worse condition than the other configurations.
To further investigate the impact of skip connections, we train the ViT-Tiny models from scratch under three configurations: the standard SAB and FFN with skip connections, SAB without skip connections, and FFN without skip connections.  

\begin{figure}[t]
    \centering
    \includegraphics[width=0.95\linewidth]{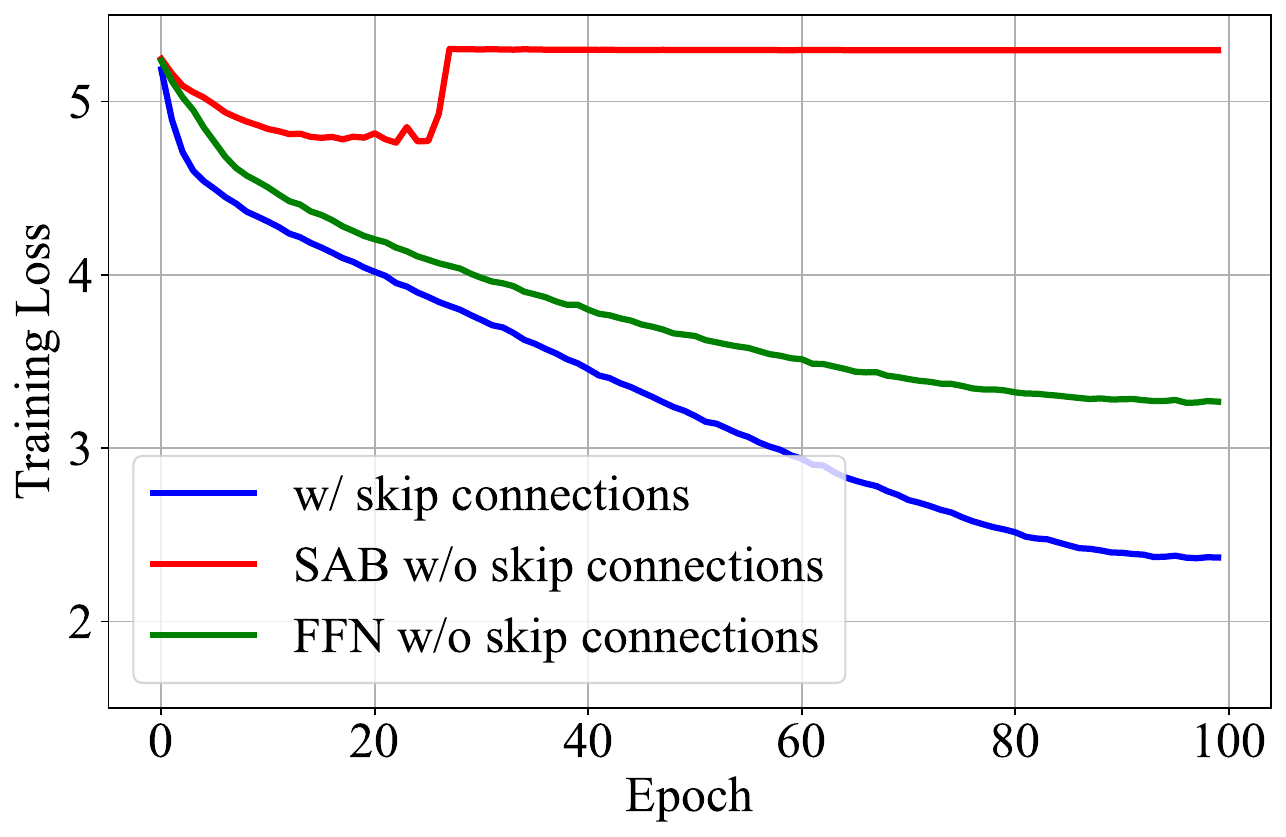}
    \caption{Training loss for three different configurations of ViT-Tiny models trained on the Tiny-ImageNet dataset.}
    \label{fig:loss}
\end{figure}

In Fig.~\ref{fig:front_fig}~(c), we observe a modest drop in classification accuracy ($2\%$) a ViT-Tiny model trained on the CIFAR-10 dataset when skip connections are removed from the FFN while retained in the SAB. However, removing skip connections from the SAB, while keeping them in the FFN, leads to a near-catastrophic performance drop of approximately $22\%$. 
Moreover, as dataset size increases, models without SAB skip connections degrade substantially in performance; on larger-scale datasets such as Tiny-ImageNet, these models fail to train altogether.
To the best of our knowledge, this empirical finding has not been reported in prior literature.
Additionally, the training loss curves for the three configurations are shown in \cref{fig:loss}.
The results demonstrate that the SAB without skip connections, which is significantly ill-conditioned, exhibits much slower convergence and diverges after 30 epochs.
Interestingly, this critical dependence on skip connections is unique to transformer architectures.

\vspace{0.1cm}
\noindent \textbf{Skip in ConvMixer.}
To further highlight the poor conditioning of the SAB, we conducted a similar experiment on a modern CNN ConvMixer~\cite{trockman2022patchesneed}. ConvMixer has a similar architecture and competitive performance to ViT with the exception that self-attention is replaced by a convolution block to spatially mix tokens. Unlike ViT, training ConvMixer-Tiny with and without skip connections, we observed that the performance remains effectively unchanged, with variations of less than ±0.2\%. Further details are shown in \cref{fig:conv} in Appendix. This contrast further suggests that the SAB introduces an extreme conditioning issue, making skip connections indispensable for stable training in ViTs.

\vspace{0.2cm}
Our main contributions are summarized as follows.
\begin{enumerate}[left=3pt, label=\arabic*.]
    \item[1. ] We present a proposition that characterizes why SAB output embedding without skip connection is fundamentally ill-conditioned, which challenges training convergence and stability (\cref{subsec4.1}).
    \item[2. ] A theoretical analysis on the role of skip connection within the SAB is undertaken. We demonstrate that it significantly improves the condition of the block's output embedding, enhancing stability and performance (\cref{subsec4.2}).
    \item[3. ] Finally, we propose a novel approach---Token Greying (TG)---to better pre-condition the input tokens. This step, when used in conjunction with conventional skip connections---improves ViT in both supervised and self-supervised settings (\cref{sec5}).
\end{enumerate}

\noindent  
Our central contribution in this paper is not TG itself, but the insight that the SAB within modern ViTs is intrinsically ill-conditioned.
Our hope is that this insight opens up brand new lines of inquiry for the vision community for more effective and efficient ViT design.

\section{Related work}
\label{sec:related_work}

\paragraph{Self-Attention.}
Originally proposed by Vaswani \etal~\cite{vaswani2017attention}, the self-attention mechanism has become central to transformer architectures, enabling the capture of long-range dependencies by dynamically weighting interactions between tokens
because it captures long-range dependencies by dynamically weighting interactions between tokens. Building on this foundation, transformer-based large language models, such as Llama \cite{dubey2024llama} and Deepseek \cite{liu2024deepseek}, have significantly advanced language understanding. In computer vision, researchers introduced Vision Transformers (ViTs) \cite{dosovitskiy2020image} for image classification, achieving superior results compared to traditional convolutional neural networks. Moreover, researchers have trained Diffusion Transformers \cite{peebles2023scalable}, replacing the commonly used U-Net backbone with a transformer that operates on latent patches and inherits the excellent scaling properties of the transformer model class. Similarly, self-attention mechanisms have been successfully applied in speech recognition and physics-informed neural networks, further demonstrating their versatility across diverse applications~\cite{dong2018speech,zhao2023pinnsformer}.


\paragraph{Skip connections.} He \etal~\cite{he2016deep} introduced skip connections in ResNet to make deep networks easier to optimize and to enhance accuracy by enabling significantly increased network depth. Since their introduction, skip connections have been extensively studied and applied across various Convolutional Neural Network (CNN) architectures. Building on these ideas, Veit~\etal~\cite{veit2016residual} offer a novel interpretation of residual networks, demonstrating that they can be viewed as a collection of multiple paths of varying lengths, which contributes to the network's flexibility and resilience. Furthermore, Hayou~\etal~\cite{hayou2021stable} addresses gradient stability in deep ResNets, noting that while skip connections alleviate the issue of vanishing gradients, they may also introduce gradient explosion at extreme depths. To counter this, they propose scaling the skip connections according to the layer index, thereby stabilizing gradients and ensuring network expressivity even in the limit of infinite depth. Recently, in Table 3 of \cite{liu2025a}, the authors reported that VGG—a model without skip connections—achieves comparable results to ResNet—a model with skip connections—on modern datasets. To the best of our knowledge, only one paper challenges the skip connection in transformer architectures. In \cite{he2023deep}, the authors attempted to train deep transformer networks without using skip connections, achieving performance comparable to standard models.
However, this approach requires five times more iterations, and the reasons why skip connections are crucial for self-attention-based transformers remain unresolved. 


\section{Preliminary}

In this section, we define the self-attention blocks used in the Vision Transformer (ViT) architecture and establish the mathematical notation for key quantities referenced in subsequent sections.
For a comprehensive overview of transformers, we refer readers to~\cite{dosovitskiy2020image}.

A Vision Transformer architecture consists of \textit{L} stacked self-attention blocks (SABs) and feedforward networks (FFNs). An identity mapping, commonly referred to as a skip connection, is applied to connect the inputs and outputs of the transformation in both SAB and FFN.

\paragraph{Self-Attention Blocks in ViT.}
Formally, given an input sequence $\mathbf{X}_{\text{in}} \in \mathbb{R}^{n \times d}$, with $n$ tokens of dimension $d$, a SAB is defined as:
\begin{equation}
    \mathbf{X}_{\text{out}} := \text{SA}(\mathbf{X}_{\text{in}}) + \mathbf{X}_{\text{in}} \text{,}
\end{equation}

\noindent where the self-attention output embedding is

\begin{equation}
    \text{SA}(\mathbf{X}_{\text{in}}) = \eta (\mathbf{QK^T})\mathbf{V} ,
    \label{eq:sa}
\end{equation}

\noindent where $\mathbf{Q}=\mathbf{X}_{\text{in}} \mathbf{W}_{\text{Q}}$, $\mathbf{K}=\mathbf{X}_{\text{in}} \mathbf{W}_{\text{K}}$ and $\mathbf{V}=\mathbf{X}_{\text{in}} \mathbf{W}_{\text{V}}$ with learnable parameters $\mathbf{W}^{\text{Q}},\mathbf{W}^{\text{K}},\text{and } \mathbf{W}^{\text{V}} \in \mathbb{R}^{d \times d}$.
$\eta (\cdot)$ is an activation function and is set by default as $\softmax$ function in \cite{dosovitskiy2020image}. Additionally, some studies \cite{saratchandran2024rethinkingsoftmaxselfattentionpolynomial,zhen2022cosformer,ramapuram2025theory} have explored the possibilities of using alternative activation functions. In particular, good performance is achieved by using linear attention with an appropriate scale.
It allows us to characterize why the SAB output embedding without a skip connection is ill-conditioned.

For general vision transformers, multiple heads SA($\mathbf{X}_\text{{in}}\text{)}_i$, where $1 \leq i \leq h$, are commonly used. Learnable matrices are divided into head numbers $h$, such that $\mathbf{W}_{\text{Q}},\mathbf{W}_{\text{K}},\text{and } \mathbf{W}_{\text{V}} \in \mathbb{R}^{h \times d \times d_h}$, where $d_h = \frac{d}{h}$. Then all outputs of each attention head are concatenated together before the skip connection.

\paragraph{Feed Forward Networks in ViT.}
Formally, given an input sequence $\mathbf{X}_{\text{in}} \in \mathbb{R}^{n \times d}$, with $n$ tokens of dimension $d$, an FFN is defined as:
\begin{equation}
    \mathbf{X}_{\text{out}} := \mathbf{W}_\text{down} (g(\mathbf{W}_\text{up}(\mathbf{X}_{\text{in}}))) + \mathbf{X}_{\text{in}} \text{,}
\end{equation}

\noindent where $\mathbf{W}_\text{up} \in \mathbb{R}^{d\times 4d}$ is up projection, $\mathbf{W}_\text{down} \in \mathbb{R}^{4d\times d}$  is down projection, and $g(\cdot)$ is an activation function.

\paragraph{Conditioning of matrices.} 
Formally, for a rectangular full rank matrix $ \mathbf{A} \in \mathbb{R}^{n \times d}$, the condition number of $\mathbf{A}$ is defined below:

\begin{equation}
    \kappa(\mathbf{A}) = \|\mathbf{A}\|_2 \|\mathbf{A}^{\dagger}\|_2 = \frac{\sigma_{\text{max}}(\mathbf{A})}{\sigma_{\text{min}}(\mathbf{A})} ,
    \label{eq:cond_num}
\end{equation}

\noindent where $\| \cdot \|_2$ is the matrix
operator norm induced by the Euclidean norm. We use the Euclidean norm in this paper. $\mathbf{A^{\dagger}}$ is the pseudo inverse of the matrix $\mathbf{A}$. $\sigma_{\text{max}}$ and $\sigma_{\text{min}}$ are the maximal and minimal singular values of $\mathbf{A}$ respectively. We will use \cref{eq:cond_num} as a metric for analysis in future sections.

\section{Theoretical analysis}\label{sec4}

In this section, we provide a theoretical analysis showing that the output embedding of the SAB without skip connections is highly ill-conditioned. \textbf{This matters as the condition of the output embedding is a proxy for the condition of the sub-block's Jacobian}. We then demonstrate that an essential function of the skip connection is to improve this conditioning.

\subsection{Self-attention is ill-conditioned}\label{subsec4.1}
We attempt to validate our claim for the simpler case of linear attention. 

\begin{proposition}
    Assume $\mathbf{X} \in \mathbb{R}^{n \times d}$,$\mathbf{W_Q}$, and $\mathbf{W_K}$ and $\mathbf{W_V} \in \mathbb{R}^{d \times d}$ have entries that are independently drawn from a distribution with zero mean. The condition number of the SAB output embedding (without skip-connection) can be expressed as:
    \begin{equation}    \kappa(\mathbf{XW_\text{Q}W_\text{K}^\text{T}X^\text{T}XW_\text{V}}) \leq C \cdot \left(\frac{\sigma_\text{max}}{\sigma_\text{min}}\right)^3 ,
    \label{eq:cond}
    \end{equation}
    \label{prop:1}
\end{proposition}
\noindent where $\sigma_\text{max}$ and $\sigma_\text{min}$ are the maximal and minimal singular values of $\mathbf{X}$. $C$ is a fixed constant that is statistically close to unity. 


Proposition \ref{prop:1} assumes the use of linear attention (\ie, no $\softmax$) as advocated for in~\cite{saratchandran2024rethinkingsoftmaxselfattentionpolynomial}. This was done to offer a simplified theoretical exposition on how the SAB without skip connection can affect the condition of the output embedding. The proposition shows the condition number of the output embedding without skip connection is bounded by the cube of the condition number of the input matrix $\mathbf{X}$. Unless the condition of $\mathbf{X}$ is unity, the output of the SAB without skip connection will therefore result in a poorly conditioned embedding. Applying this same process across multiple layers within a transformer results in a highly ill-conditioned outcome.

\begin{figure}[t!]
\vspace{1em}
    \centering
    \begin{subfigure}[t]{\linewidth}
        \centering
        \includegraphics[width=\linewidth]{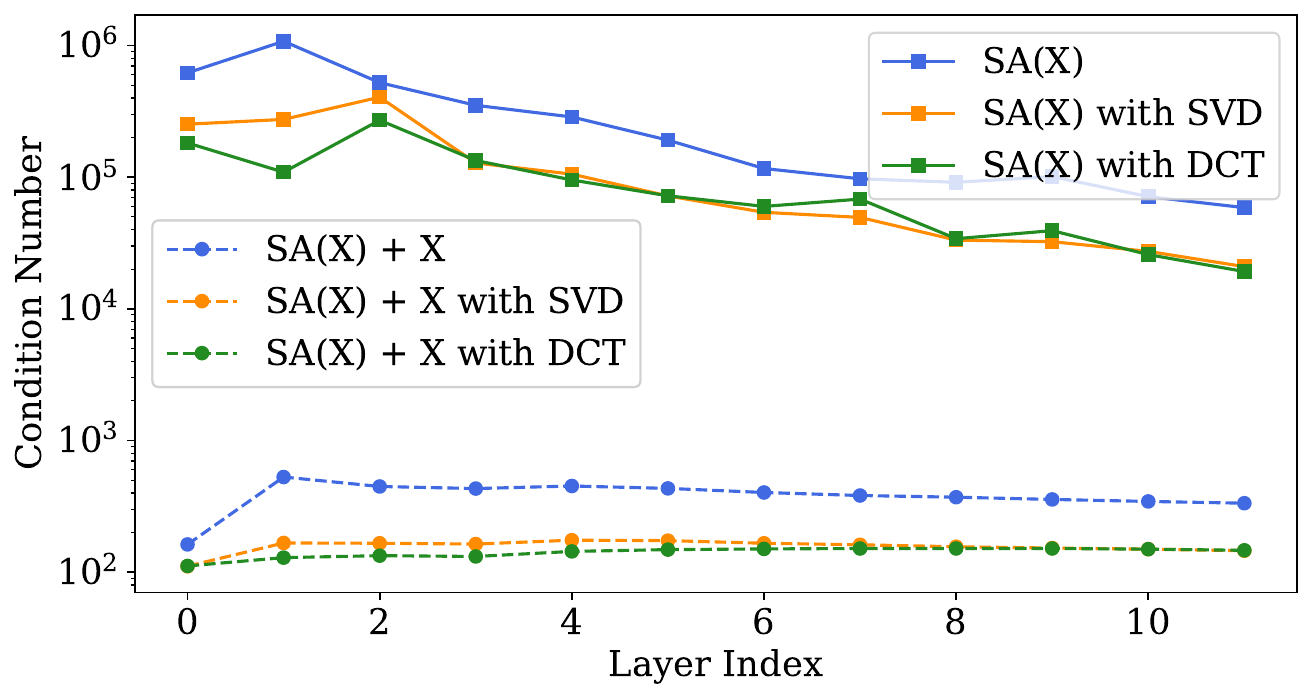}  
    \end{subfigure}
    \begin{subfigure}[t]{\linewidth}
        \vspace{-0.1cm} 
        \centering
        \includegraphics[width=\linewidth]{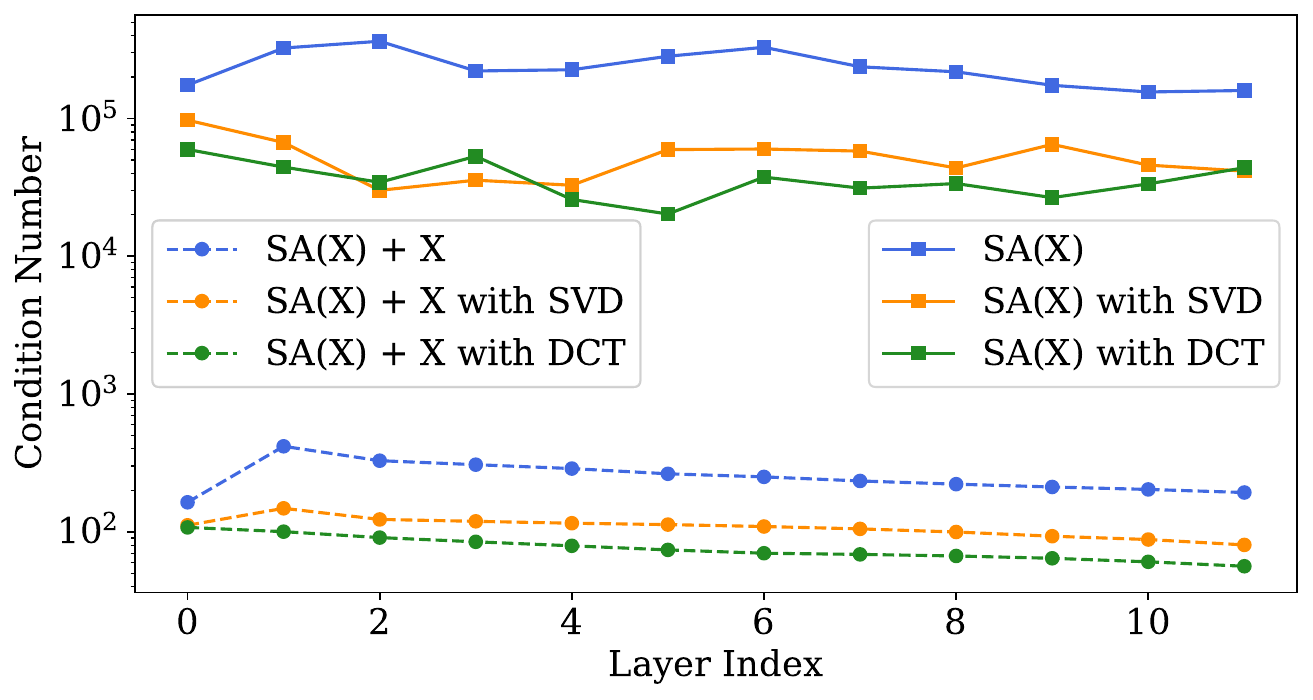}  
    \end{subfigure}

    \caption{Condition numbers across all layers of trained ViT-Tiny on Tiny-ImageNet datasets. We use 10\% validation sets to plot the above figures. \textbf{Upper: }fixed scale linear self-attention. \textbf{Lower: }$\softmax$ self-attention}
    \label{fig:linear_softmax_condition_number}
\end{figure}

\cref{fig:linear_softmax_condition_number} illustrates our theoretical analysis by showing that the linear self-attention operation increases the condition number of embeddings from $e^3$ to $e^6$, and we can empirically extend this analysis to \textit{softmax} self-attention. The output embedding of the \textit{softmax} SAB without skip connections is also significantly ill-conditioned.

Further, for condition numbers of FFN output embeddings without skip connection using linear activation $g$ we have:
\begin{equation}
    \kappa (\text{MLP}(\mathbf{X})) = \kappa(\mathbf{W_\text{down}W_\text{up}X}) \leq C_\text{down}\cdot C_\text{up} \cdot\frac{\sigma_\text{max}}{\sigma_\text{min}} ,
\end{equation}

\noindent where $\sigma_\text{max}$ and $\sigma_\text{min}$ are maximal singular values of $\mathbf{X}$ computed using SVD. Since $\mathbf{W}_\text{down}$ and $\mathbf{W}_\text{up}$ do not depend on data $\mathbf{X}$, we denote $\kappa(\mathbf{W_\text{down}}) = C_\text{down}$ and $\kappa(\mathbf{W_\text{up}}) = C_\text{up}$.

Compared to the self-attention mechanism, the MLP transformation does not worsen conditioning as severely, since it has a much lower bound on the condition number. In Fig.~\ref{fig:front_fig} (b), the output embeddings of the FFN without skip connection is observed to have a significantly better condition than the output embedding of the SAB without skip connection. Therefore, in (c), the performance drop after removing the FFN's skip connections is less severe than after removing the SAB's skip connections.

\subsection{Skip connection improves condition}\label{subsec4.2}
In this subsection, we theoretically demonstrate that the skip connection improves the condition of the self-attention output embeddings.

Previous works~\cite{he2023deep, noci2024shaped, dong2021attention} have demonstrated that transformers, without skip connections, converge very slowly or may even become un-trainable.
In \cite{he2023deep}, the authors show that without skip connections, output embeddings of the SAB quickly converge to \text{rank 1}. They hypothesize that this is due to the inability of the transformer to propagate signals through the self-attention mechanism. This rank-1 output embedding implies an infinite condition number. The authors introduce an inductive bias into the self-attention mechanism to enable what they refer to as ``better signal propagation''. While their approach matches the performance of standard (vanilla) models, it incurs a substantial computational cost—running approximately five times slower.
%
We hypothesize that the success of this method could be directly attributed to its ability to improve the condition of the self-attention output embeddings. 

This observation further motivates our view that the primary role of skip connections may be to regularize the condition of the self-attention output embeddings

\begin{proposition}
    \label{condition_theorem}
Let $\mathbf{X} \in \mathbb{R}^{n \times d}$ and $\mathbf{W_{Q}}, \mathbf{W_{K}},\mathbf{W_V} \in \mathbb{R}^{d \times d}$ have entries that are independently drawn from a distribution with zero mean. We define \mbox{$\mathbf{M} = \mathbf{W_Q W^T_K X^T X W_V} $} and assume $\mathbf{M}$ is the positive semi-definite matrix. We have the following bounds on the condition numbers $\kappa$.
\begin{equation}
    \kappa(\mathbf{X M+X}) \ll  \kappa(\mathbf{X M}).
    \label{eq:self_attention_cond}
\end{equation}
\end{proposition}

\cref{condition_theorem} shows that during training, the condition number of the SAB output embedding has a much lower bound than that of the SAB output embedding without skip connections. 
Furthermore, with other conditions being equal, better condition of SAB can contribute to faster convergence and more stable gradient updates, reducing the risk of gradient explosion or vanishing. 
This ensures the signal can propagate well during the training process in the Vision Transformers. 

In other words, the self-attention mechanism enables tokens to communicate by focusing on different parts of the input, thereby capturing correlations between elements within the same sequence. However, one critical drawback is that it is ill-conditioned and challenging to train. In contrast, although skip connections do not facilitate communication between tokens, they help regularize the condition of output embeddings and aid in the optimization process.

As shown in~\cref{fig:linear_softmax_condition_number}, empirical results demonstrate that in both the linear and $\softmax$ self-attention cases, adding skip connections significantly improves the condition of output embeddings.

\section{Methodology}\label{sec5}

In this section, we introduce \textbf{T}oken \textbf{G}raying \textbf{(TG)}, a simple yet effective method that employs Singular Value Decomposition and Discrete Cosine Transform techniques to reconstruct better-conditioned input tokens.

\begin{algorithm}[t!]
\centering
\caption{SVD Token Graying}
\label{alg1}

    \begin{algorithmic}[1]
    \State \(\triangleright\) \textbf{Input:} training data, amplification coefficient $\epsilon \in (0,1]$
    \State \(\triangleright\) \textbf{Training:}
    \For{each minibatch \(\{(\mathbf{x}_i, y_i)\}_{i=1}^k \)}
        \State Convert image \(\mathbf{x}_i\) into token  \(\mathbf{X}\)
        \State $\mathbf{ U\Sigma V^T \leftarrow  X}$ \Comment{Compute SVD}
        \State $\mathbf{\Tilde{\Sigma} \leftarrow  \frac{\Sigma}{\text{max}(\Sigma)}}$ \Comment{Normalize elements $\in (0,1]$}
        \State $\mathbf{\Tilde{\Sigma} \leftarrow \Tilde{\Sigma}^{\epsilon}}$ \Comment{Amplify}
        \State $\mathbf{\Tilde{X} \leftarrow U\Tilde{\Sigma} V^T}$ \Comment{SVD  reconstruct}
        \State Patch Embedding: \( \mathbf{Z} \leftarrow \text{PatchEmbed}(\tilde{\mathbf{X}}) \)
        \State Forward Pass: \( \hat{y} \leftarrow \text{Model}(\mathbf{Z}) \)
        \State Compute Loss: \( \mathcal{L} \leftarrow \text{Loss}(\hat{y}, y_i) \)
        \State Backward Pass: Compute gradients \(\nabla_{\theta} \mathcal{L}\)
        \State Parameter Update: \(\theta \leftarrow \theta - \eta \nabla_{\theta} \mathcal{L}\)
    \EndFor
    \end{algorithmic}

\end{algorithm}

\subsection{Singular Value Decomposition (SVD)}
Singular Value Decomposition of a matrix $\mathbf{X} \,{\in}\, \mathbb{R}^{n {\times} d}$ is the factorization of $\mathbf{X}$ into the
product of three matrices $\mathbf{X=U \Sigma V^T}$ where the columns of $\mathbf{U} \,{\in}\, \mathbb{R}^{n {\times} n}$ and $\mathbf{V} \,{\in}\, \mathbb{R}^{d {\times} d}$ are orthonormal and
$\mathbf{\Sigma} \,{\in}\, \mathbb{R}^{n {\times} d}$ is a rectangular diagonal matrix whose diagonal elements are non-negative in descending order and represent the singular values such that \mbox{$\Sigma_{ii} \,{=}\, \sigma_i$}. As observed in \cref{eq:cond_num}, we can reduce the condition number by increasing the minimal singular value while keeping the maximal singular value unchanged. Hence, our goal is to reconstruct $\mathbf{\Tilde{X}} = \mathbf{U\tilde{\Sigma}}\mathbf{V^T}$, where $\mathbf{\tilde{\Sigma}}$ is formed by amplifying non-maximal elements of $\mathbf{\Sigma}$ while keeping the maximal element unchanged. Our SVD token graying pipeline is presented in \cref{alg1}.

However, directly applying SVD to matrices is computationally expensive, with a cost of \mbox{$O(nd \cdot \text{min}(n,d))$} leading to longer training time. \cref{tab:training_time} demonstrates that training a ViT is significantly slower when using SVD reconstruction. Therefore, in the next subsection, we introduce a more efficient \textbf{T}oken \textbf{G}raying (\textbf{TG}) method using the discrete cosine transform. 

\begin{table}[ht!]
    \centering
    \scalebox{0.9}{
    \begin{tabular}{l|c|c|c}
    \toprule
         Methods & ViT-B & ViT-B + SVDTG & ViT-B + DCTTG \\
         \midrule 
        Time (days) &0.723 & 4.552 & 0.732  \\
        \bottomrule
    \end{tabular}}
    
    \caption{ViT-Base training time w/o and w/ token graying methods on ImageNet-1K dataset.}
        \label{tab:training_time}
\end{table}

\subsection{Discrete Cosine Transform (DCT)}

A Discrete Cosine Transform (DCT) expresses a finite sequence of data points as a sum of cosine functions oscillating at different frequencies. It is real-valued and has an inverse, which is often simply called the Inverse DCT (IDCT). 


There are several variants of the DCT and in this work we choose DCT-II which is one of the most commonly used forms. Given a real-valued sequence data $\mathbf{x} \in \mathbb{R}^{N \times 1}$, the 1D DCT sequence $\hat{\mathbf{x}}$ is defined as:

\begin{equation}
    \begin{aligned}
    \hat{\mathbf{x}}_k = \alpha_k \sum_{i=0}^{N-1} & x_i \cos \left[ \frac{\pi(2i+1)k}{2N}\right], \text{for } k = 0, 1, ..., N-1 \\
    &\text{where } \alpha_k = 
    \begin{cases}
            {\sqrt{\frac{1}{N}}},  \text{if } k = 0 \\
            {\sqrt{\frac{2}{N}}},  \text{if } k \neq 0 \\
    \end{cases}
\end{aligned}
\label{eq:dct}
\end{equation}

Simplicitly, 1D DCT is a linear transformation,~\cref{eq:dct} can be written with the formula $DCT(\mathbf{x}) \,{=}\, \mathbf{\hat{x}} \,{=}\, D\mathbf{x}$, where $D_{i,k} $ is the 1D DCT basis and orthogonal. 

\begin{align}
    D_{i,k} = \alpha_k \cos \left[ \frac{\pi(2i+1)k}{2N}\right].
\end{align}

For the reconstruction process, since the DCT expresses a signal as a sum of cosine functions with different frequencies, the IDCT combines these frequency components to recover the original signal. Hence, IDCT$(\hat{x}) \,{=}\, \Tilde{\mathbf{x}}=D^T \mathbf{\hat{x}}$. In our case, we have a token matrix $\mathbf{X} \,{\in}\, \mathbb{R}^{n {\times} d}$ and we generalize the 1D DCT to a 2D DCT using the formula \mbox{$\mathbf{\hat{X}} \,{=}\, D \mathbf{X}D^T$}. 
In natural images, the dominant singular vector, associated with the largest singular value, typically captures the low frequency content of the image.
Intuitively, this observation suggests that DCT reconstruction can serve as an approximation for SVD reconstruction.
Our DCT token graying pipeline is presented in \cref{alg2}.

DCT avoids the expensive computational cost of SVD and has a complexity of \mbox{$O(nd \cdot \log (nd))$}. \cref{tab:training_time} shows that training ViT using the DCT algorithm is significantly faster than SVD algorithm.

\begin{algorithm}[t]
\caption{DCT Token Graying}
\label{alg2}
    \begin{algorithmic}[1]
    \State \(\triangleright\) \textbf{Input:} training data, amplification coefficient $\epsilon \in (0,1]$
    \State \(\triangleright\) \textbf{Training:}
    \For{each minibatch \(\{(\mathbf{x}_i, y_i)\}_{i=1}^k \sim \text{training data}\)}
        \State Convert image \(\mathbf{x}_i\) into token  \(\mathbf{X}\)
        \State $ D\mathbf{\hat{X}}D^T \leftarrow \mathbf{X}$ \Comment{Compute DCT}
        \State $\mathbf{\hat{X}}_\text{norm} \leftarrow  |\mathbf{\hat{X}}|/\text{max}(|\mathbf{\hat{X}}|)$ \Comment{Normalize elements $\in (0,1]$}
        \State $\mathbf{\hat{X}} \leftarrow \mathbf{\hat{X}}_\text{norm}^{\epsilon}\cdot \text{sign}(\mathbf{\hat{X}})\cdot\text{max}(|\mathbf{\hat{X}}|) $ \Comment{Amplify}
        \State $\mathbf{\Tilde{X}} \leftarrow \hat{D}^T \mathbf{\hat{X}}\hat{D}$ \Comment{Inverse DCT}
        \State Patch Embedding: \( \mathbf{Z} \leftarrow \text{PatchEmbed}(\tilde{\mathbf{X}}) \)
        \State Forward Pass: \( \hat{y} \leftarrow \text{Model}(\mathbf{Z}) \)
        \State Compute Loss: \( \mathcal{L} \leftarrow \text{Loss}(\hat{y}, y_i) \)
        \State Backward Pass: Compute gradients \(\nabla_{\theta} \mathcal{L}\)
        \State Parameter Update:  \(\theta \leftarrow \theta - \eta \nabla_{\theta} \mathcal{L}\)
    \EndFor
    \end{algorithmic}
\end{algorithm}

\section{Experiments}\label{sec6}
Vision Transformers (ViTs) have emerged as powerful models in the field of computer vision, demonstrating remarkable performance across a variety of tasks. 
This section is dedicated to validating and analyzing our proposed algorithm for both supervised and self-supervised learning in ViTs.
For all experiments, unless specified otherwise, we use $\epsilon \,{=}\, 0.95$.
\subsection{Supervised learning ViT}
In this subsection, we validate our methods in supervised learning using different types and scales of ViTs.
\textit{Cross-entropy} loss objective is used.
We use PyTorch and Timm libraries.

\vspace{0.2cm}
\noindent \textbf{-- ViT} \cite{dosovitskiy2020image}: 
is the pioneering work that interprets an image as a sequence of patches and processes it by a standard Transformer encoder as used in NLP. This simple, yet scalable, strategy works surprisingly well when coupled with pre-training on large datasets.
We train the ViT-Tiny, which has 12 layers and 3 heads, with a head dimension of 64 and a token dimension of 192 on the Tiny-ImageNet  dataset~\cite{le2015tiny}. Additionally, we train ViT-Base, which consists of 12 layers and 12 heads, each with a head dimension of 64 and a token dimension of 768 on the ImageNet-1K dataset~\cite{deng2009imagenet}.

\vspace{0.2cm}
\noindent \textbf{-- Swin} \cite{liu2021swin} is a hierarchical vision transformer designed to focus on local attention first and expand globally. It improves computational efficiency and locality modeling compared to standard ViTs. We train on the Swin small model with a patch size of 4 and a window size of 7 on the ImageNet-1K dataset.

\vspace{0.2cm}
\noindent \textbf{-- CaiT} \cite{touvron2021going} is a variant Vision Transformer that aims at increasing the stability of the optimization when ViTs go deeper. We train CaiT small, which has 24 layers, on the ImageNet-1K dataset.

\vspace{0.2cm}
\noindent \textbf{-- PVT} \cite{wang2021pyramid} an enhanced version of the Pyramid Vision Transformer that utilizes a hierarchical architecture to extract multi-scale features for improved performance in computer vision tasks. It incorporates efficient attention mechanisms and refined positional encoding to boost accuracy and reduce computational complexity.

\begin{table}[t]
    \centering
    \vspace{1em}
    \begin{tabular}{l c c}
        \toprule
        Method & Top-1 Acc (\%) & Top-5 Acc (\%)  \\
        \midrule
        $\frac{1}{k}(\mathbf{QK})$        & 43.0 &  62.7\\
        $\frac{1}{k}(\mathbf{QK})$ + SVD & 44.3 &  68.8 \\
        $\frac{1}{k}(\mathbf{QK})$ + DCT & 44.2 &  68.8 \\
        \midrule
        $\softmax (\mathbf{QK})$ & 43.0  & 67.3 \\
        $\softmax (\mathbf{QK})$ + SVD & 44.7 & 69.7 \\
        $\softmax (\mathbf{QK})$ + DCT & 44.8  & 69.7 \\
        \bottomrule
    \end{tabular}
    \caption{ViT-Tiny results on Tiny-ImageNet datasets.  In the upper part, we show results for linear self-attention, where $k= 16$ is chosen based on \cite{saratchandran2024rethinkingsoftmaxselfattentionpolynomial} w/ and w/o our method. On the lower part, we show regular \textit{softmax} self-attention results w/ and w/o our method.}
    \label{vit_tiny_tiny}
\end{table}

\begin{table}[t]
    \centering
    \scalebox{0.9}{
    \begin{tabular}{l|c|c|c|c}
        \toprule
        Method & $\epsilon$ & Top-1 Acc (\%) & $\kappa_\text{in}$ &$\kappa_\text{out}$ \\
        \midrule
        ViT-Base & -- & 81.0 & 6.72 &6.74 \\
        \midrule
        \multirow{5}{*}{ViT-Base + SVDTG} & 0.9 & 81.2 & 6.64 & 6.66\\
        & 0.8 & 81.2 &6.47 & 6.66\\
        & 0.7 & \textbf{81.4} &6.15 &6.17\\
        & 0.6 & \textbf{81.4} &5.73&5.71\\
        & 0.5 & 81.0 &5.29&5.25\\
        \midrule
        \multirow{5}{*}{ViT-Base + DCTTG} & 0.97 & 81.1 &6.47 &6.49\\
        & 0.95 & \textbf{81.3} &  6.42&6.43\\
        & 0.93 & 81.2 & 6.33&6.35\\
        & 0.9 & 81.1 &6.25&6.25\\
        & 0.85 & 81.0& 6.01 &6.02\\
        \bottomrule
    \end{tabular}}
    \caption{ViT-Base results using SVD and DCT token graying methods with different amplification coefficient $\epsilon$. $\kappa_\text{in}$ and $\kappa_\text{out}$ are the average token condition numbers in log scale before and after each SAB.}
    \label{vitb_svd_dct}
    \vspace{-0.2cm} 
\end{table}

\begin{figure}[ht]
    \centering
    \centering
    \begin{subfigure}[t]{\linewidth}
        \centering
        \includegraphics[width=\linewidth]{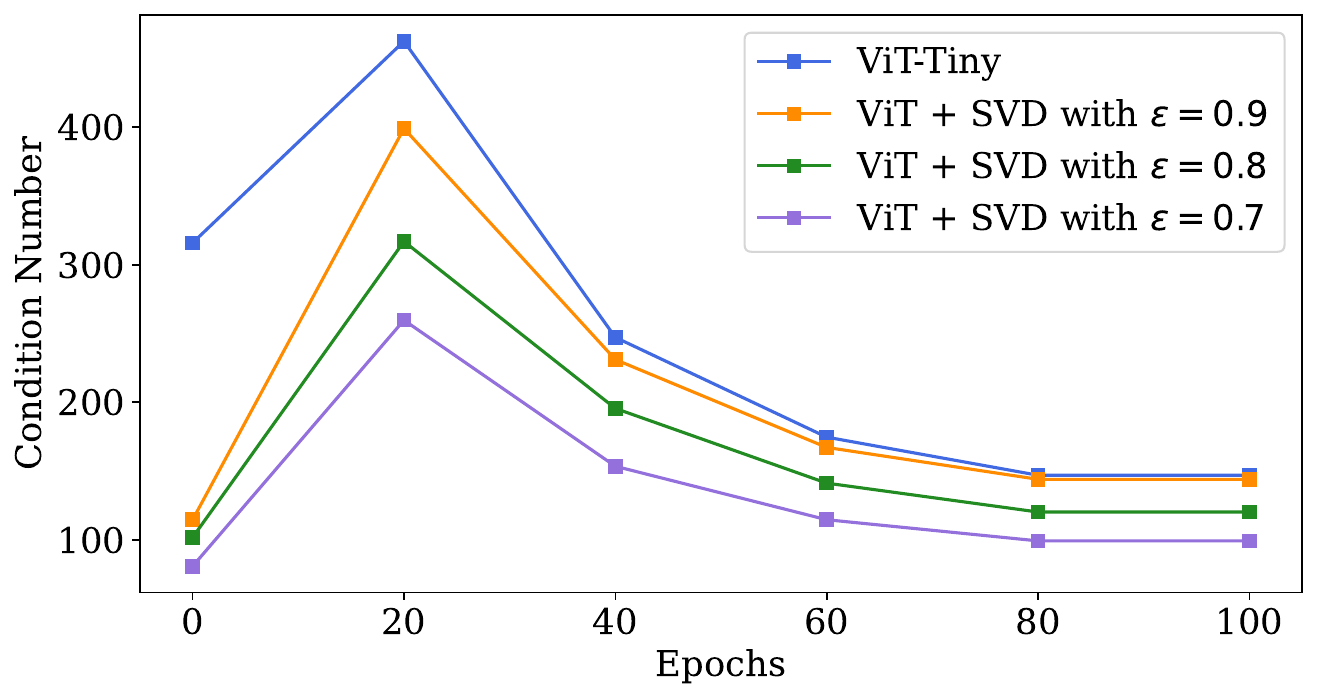}  
    \end{subfigure}
    \begin{subfigure}[t]{\linewidth}
        \centering
        \includegraphics[width=\linewidth]{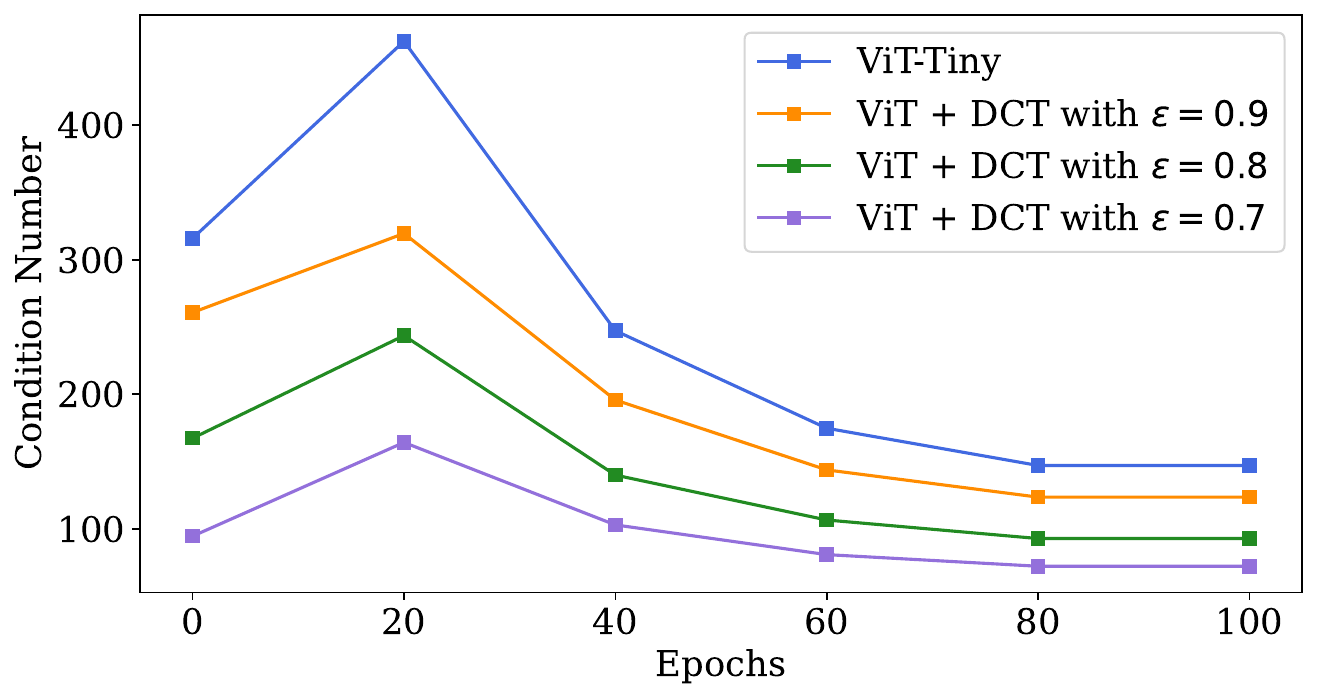}  
    \end{subfigure}

    \caption{We compute the condition number of the SAB block's output embeddings during ViT-Tiny training using SVDTG and DCTTG with different $\epsilon$.}
    \label{fig:training_condition_numbers}
\end{figure}

\begin{figure}[t]
\vspace{0.2cm}

    \centering
    \includegraphics[width=1\linewidth]{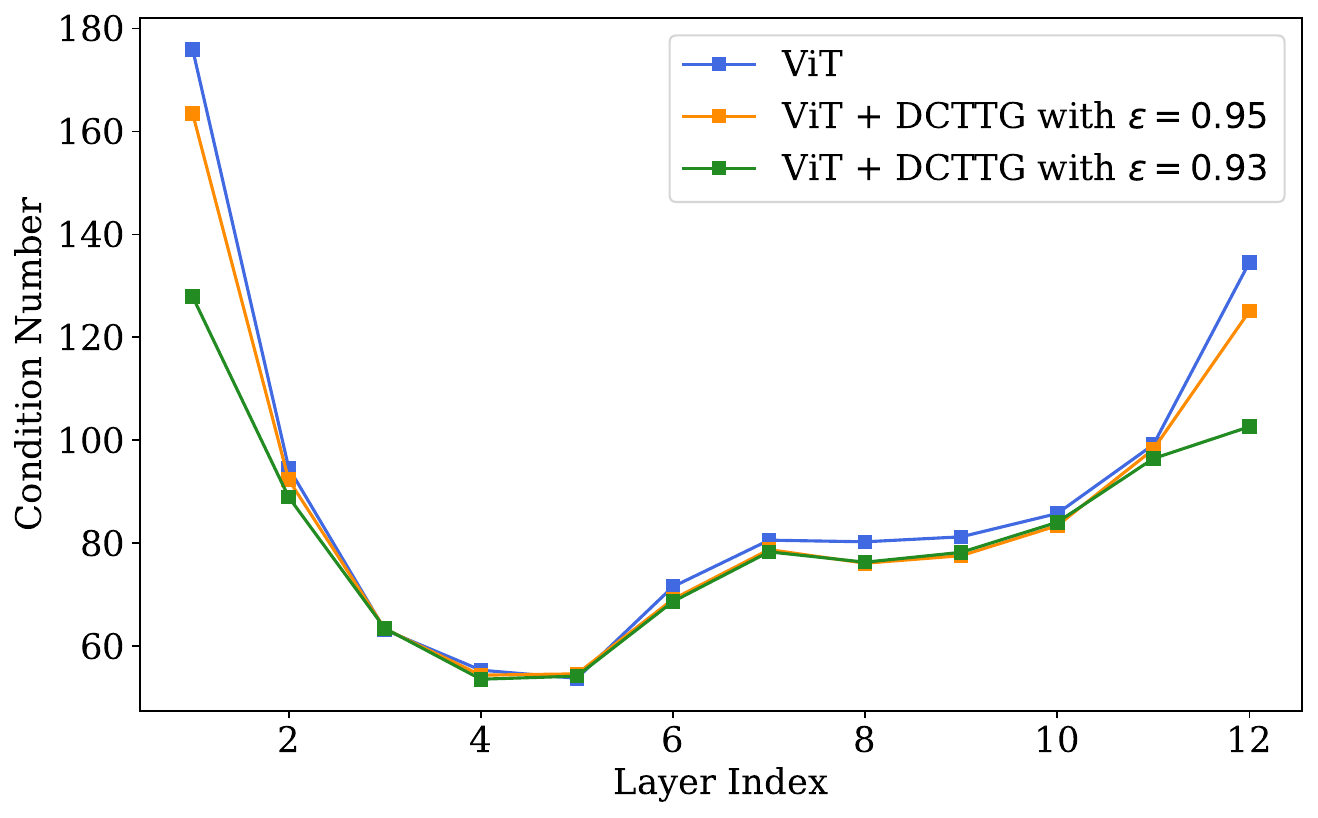}
    \caption{Condition numbers of all layers in the trained ViT-Base model with and without the DCTTG method.}
    \label{fig:vitb_condition_all layer}
\end{figure}

\vspace{0.2cm}
\noindent \textbf{Results. } In \cref{vit_tiny_tiny}, we first demonstrate the results for the linear case and the regular $\softmax$ case using our methods on a small model (ViT-Tiny) and the small-scale Tiny-ImageNet dataset. Given that Theorem \cref{condition_theorem} and Proposition \cref{prop:1} are mathematically explained using linear self-attention, we train the ViT-Tiny model with linear self-attention, which aligns well with our analysis in \cref{fig:linear_softmax_condition_number}. Furthermore, as shown in \cref{fig:training_condition_numbers}, both SVDTG and DCTTG lead to better conditioning of SAB output embeddings throughout training epochs. 
Next, we train the model on a larger dataset using the default $\softmax$ self-attention, further validating that our approach generalizes to $\softmax$ self-attention. 
Moreover, SVDTG is the most direct method to improve the conditioning of a matrix, and as demonstrated in \cref{vitb_svd_dct}, it leads to notable performance gains. 
To address the high computational cost---approximately six times slower than regular ViT training---we approximate SVDTG using DCTTG, which achieves better token embeddings conditioning and yields comparable results too. 
As shown in~\cref{fig:vitb_condition_all layer}, our DCTTG method produces condition numbers across all layers that indicate better token conditioning in almost every layer.
Finally, we evaluate our methods on different variants of ViTs trained on ImageNet-1K datasets, where our DCTTG method demonstrates superior performance compared to vanilla models in terms of Top-1 and Top-5 classification accuracy, while also achieving better conditioning of SAB output embeddings across all layers in the ViT-B model.

\begin{table}[t]
    \centering
    \begin{tabular}{l|c|c}
            \toprule
        Method & Top-1 Acc (\%) & Top-5 Acc (\%) \\
        \midrule
        ViT-S & 80.2 & 95.1 \\
        ViT-S + DCTTG & \textbf{80.4} & \textbf{95.2} \\
        \midrule
        ViT-B & 81.0 & 95.3 \\
        ViT-B + DCTTG & \textbf{81.3} & \textbf{95.4} \\
        \midrule
        Swin-S & 81.3 & \textbf{95.6}  \\
        Swin-S + DCTTG & \textbf{81.6} & \textbf{95.6} \\
        \midrule
        CaiT-S & 82.6 & 96.1 \\
        CaiT-S + DCTTG & \textbf{82.7} & \textbf{96.3} \\
        \midrule
        PVT V2 b3 & 82.9 & 96.0 \\
        PVT V2 b3 + DCTTG & \textbf{83.0} & \textbf{96.1}  \\
        \bottomrule
    \end{tabular}
    \caption{Top-1 and Top-5 classification accuracy on ImageNet-1K dataset using DCT token graying on different ViTs configurations.}
    \label{tab:my_label}
    \vspace{-0.5cm}
\end{table}

\subsection{Self-supervised learning ViT}
In this subsection, we validate our methods on self-supervised learning models as well as on fine-tuning pretrained models~\cite{he2022masked, haghighat2024ropim}. Self-supervised learning is a powerful technique in which models learn representations directly from unlabeled data.
Masked Image Modeling (MIM) \cite{he2022masked} learns representations by reconstructing images that have been corrupted through masking.

\vspace{0.2cm}
\noindent \textbf{Pre-training. } 
In the pre-training stage, we pre-train our model on the ImageNet-1k training set without using ground-truth labels in a self-supervised manner. 
We use a batch size of $2048$, a base learning rate of $1.5 \times 10^{-4}$, and a weight decay of $0.05$, and we train for $400$ epochs. 
Our loss function computes the mean squared error (MSE) between the reconstructed and original images in pixel space.

\vspace{0.2cm}
\noindent \textbf{Finetuning and reuslts. } 
In the finetuning stage, we fine-tune the pre-trained models on ImageNet-1k for classification tasks using \textit{cross-entropy} loss. 
We use a batch size of $512$, a base learning rate of $5 \times 10^{-4}$, a weight decay of $0.05$, and train for 100 epochs. As shown in \cref{mae_vit_base_finetune}, our DCTTG method outperforms the vanilla MAE model by 0.2\% in both top-1 and top-5 classification accuracy.
\begin{table}[t]
    \centering
    \begin{tabular}{ l l l}
        \toprule
          Method & Top-1 Acc (\%) & Top-5 Acc (\%) \\
         \midrule
MAE & 83.0  & 96.4  \\
         MAE + DCTTG & \textbf{83.2} & \textbf{96.6} \\
        \bottomrule
    \end{tabular}
    \caption{MAE pretrained ViT-Base results finetuned on ImageNet dataset-1k. We evaluate our MSVR method on self-attention using the power iteration method.}
    \label{mae_vit_base_finetune}
    \vspace{-0.2cm}
\end{table}

\section{Limitation}
Our work establishes a theoretical foundation demonstrating why skip connections act as essential conditioning regularizers within self-attention blocks. Additionally, we introduce a method designed to further enhance the output embeddings of self-attention blocks. However, certain limitations remain. For instance, while our regularizer is effective, training in low-precision settings may pose challenges due to the DCT operation, which involves many multiplications and summations that are sensitive to quantization errors. Additionally, for some variants of ViTs, the performance improvement might be marginal.

\section{Conclusion}
This paper provides a theoretical demonstration that the output embeddings of linear self-attention blocks, when lacking skip connections, are inherently ill-conditioned, a finding that empirically extends to conventional $\softmax$ self-attention blocks as well. Furthermore, we show that skip connections serve as a powerful regularizer by significantly improving the conditioning of self-attention blocks' output embeddings. In addition, we use the condition of the self-attention output embeddings as a proxy for that of its Jacobian, empirically demonstrating that input tokens with improved conditioning lead to a better-conditioned Jacobian.
Building on this insight, we introduce SVDTG and DCTTG, simple yet effective methods designed to improve the condition of input tokens. We then validate their effectiveness on both supervised and unsupervised learning tasks. Ultimately, we hope these insights will guide the vision community in designing more effective and efficient ViT architectures for downstream applications.

%
\vspace{0.2cm}
\noindent\textbf{Acknowledgements.} This work was supported by the Australian Institute for Machine Learning (AIML) and the CSIRO’s Data61 Embodied AI Cluster. 


{
    \small
    \bibliographystyle{ieeenat_fullname}
    \bibliography{main}
}

\clearpage
\maketitlesupplementary
\appendix

\section{Theoretical framework}
In this section, we give the proof of our \cref{prop:1}.

\begin{proof}

By the properties of 2-norm condition number $\kappa(\mathbf{AB}) \leq \kappa(\mathbf{A})\cdot\kappa(\mathbf{B})$, then we can rewrite left sides of Eq.\eqref{eq:cond} as follows:

\begin{equation}
    \kappa(\mathbf{XW_\text{Q}W_\text{K}^\text{T}X^\text{T}XW_\text{V}}) 
    \leq \kappa(\mathbf{XW_\text{Q}})\cdot\kappa(\mathbf{XW_\text{K}})\cdot\kappa(\mathbf{XW_\text{V}}).
\end{equation}

\noindent By definition we have:
\begin{equation}
    \kappa(\mathbf{XW_\text{Q}}) \leq \kappa(\mathbf{X}) \cdot \kappa(\mathbf{W_\text{Q}}).\\
\end{equation}
Since $\mathbf{W}_\text{Q}$ does not depend on data $\mathbf{X}$, we denote $\kappa(\mathbf{W_\text{Q}}) = C_\text{Q}$.
Then we have 
\begin{equation}
    \kappa(\mathbf{XW_\text{Q}}) \leq C_\text{Q} \cdot \kappa(\mathbf{X}). \\
\end{equation}

\noindent Similarly, this holds for $\kappa(\mathbf{XW_\text{K}})$ and $\kappa(\mathbf{XW_\text{V}})$ as well. Since all weight matrices stem from a zero mean i.i.d. with high statistical likelihood $C = C_{Q} \cdot C_{K} \cdot C_{V}$ will tend towards unity. 
This completes the proof.
\end{proof}

\noindent Next, we give the proof of our \cref{condition_theorem}

\begin{proof}
     By the properties of 2-norm condition number $\kappa(\mathbf{AB}) \leq \kappa(\mathbf{A})\cdot\kappa(\mathbf{B})$, then we can rewrite left sides of Eq.\eqref{eq:self_attention_cond} as follows:

\begin{align}
      &\kappa(\mathbf{X M+X})  \\
    = &\kappa(\mathbf{X (M+I)}) \\
    \leq &\kappa(\mathbf{X})\cdot \kappa(\mathbf{M+I}) .
\end{align}
Then we take the notation $C = C_{Q} \cdot C_{K} \cdot C_{V}$ and we have

\begin{align}
    \kappa(\mathbf{M}) &\leq \frac{C \cdot \sigma_\text{max}^2}{\sigma_\text{min}^2} \\
    \kappa(\mathbf{M+I}) &\leq\frac{C \cdot \sigma_\text{max}^2 + 1}{\sigma_\text{min}^2 + 1}
\end{align}

\noindent Where $C = C_{Q} \cdot C_{K} \cdot C_{V}$ $\sigma_\text{max}$ and $\sigma_\text{min}$ represents maximal and minimal singular value of $\mathbf{X}$. Then we have

\begin{align}
    \kappa(\mathbf{X})\cdot\kappa(\mathbf{M}) &\leq \frac{C \cdot \sigma_\text{max}^3}{\sigma_\text{min}^3} \\
    \kappa(\mathbf{X})\cdot\kappa(\mathbf{M+I}) &\leq\frac{C \cdot \sigma_\text{max}^3 + \sigma_\text{max}}{\sigma_\text{min}^3 + \sigma_\text{min}}
\end{align}

    For a matrix $\mathbf{X} \in \mathbb{R}^{n \times d}$ whose entries are i.i.d with mean zero, it holds with high probability $\sigma_\text{min} < 1$ and $\sigma_\text{max} > 1$. Then we have 

\begin{align}
    C \cdot \sigma_\text{max}^3 + \sigma_\text{max} & \approx C \cdot \sigma_\text{max}^3 \\
    \sigma_\text{min}^3 + \sigma_\text{min} & > \sigma_\text{min}^3 
\end{align}

Therefore,
$\kappa(\mathbf{X})\cdot\kappa(\mathbf{M}+I) \ll \kappa(\mathbf{X})\cdot\kappa(\mathbf{M})$ 





This completes our proof for \cref{condition_theorem}.

\end{proof}
\section{ConvMixer}

In this section, we demonstrate how removing skip connections in Convolutional Neural Networks (CNNs) impacts performance. ConvMixer, an extremely simple model inspired by Vision Transformers (ViTs), was introduced in \cite{trockman2022patchesneed} and remains widely used within the research community.
ConvMixer consists of LL ConvMixer blocks, formally defined as follows:

\begin{align}
    \mathbf{X}_l^{'} &= \text{BN}(\sigma \{ \text{ConvDepthwise}(\mathbf{X}_{l})\} + \mathbf{X}_{l} \\
    \mathbf{X}_{l+1} &= \text{BN}(\sigma \{ \text{ConvPointwise}(\mathbf{X}_l^{'})\}
\end{align}

\noindent where $\mathbf{X} \in \mathbb{R}^{h \times n/p \times n/p}$, BN is batch normalization, and $\sigma$ is activation function. $h$ is feature dimension, $n$ is image height and width and $p$ is patch size. 

For the ConvDepthwise transformation, we analyze the condition number in the linear case and in the absence of batch normalization. We have:
\begin{equation}
    \kappa(\mathbf{W}_\text{CD}\mathbf{X}_{l}) \leq \kappa(\mathbf{W}_\text{CD})\cdot\kappa(\mathbf{X}_{l}) \leq C_\text{CD} \cdot \kappa(\mathbf{X}_{l})
\end{equation}

\noindent Since $\mathbf{W}_\text{CD}$ does not depend on data, we denote $\kappa(\mathbf{W}_\text{CD})=C_\text{CD}$. Compared to the Self-Attention Mechanism, the ConvDepthwise transformation demonstrates better conditioning with a lower bound on condition numbers.

\begin{figure}[t!]
    \centering
    \includegraphics[width=1\linewidth]{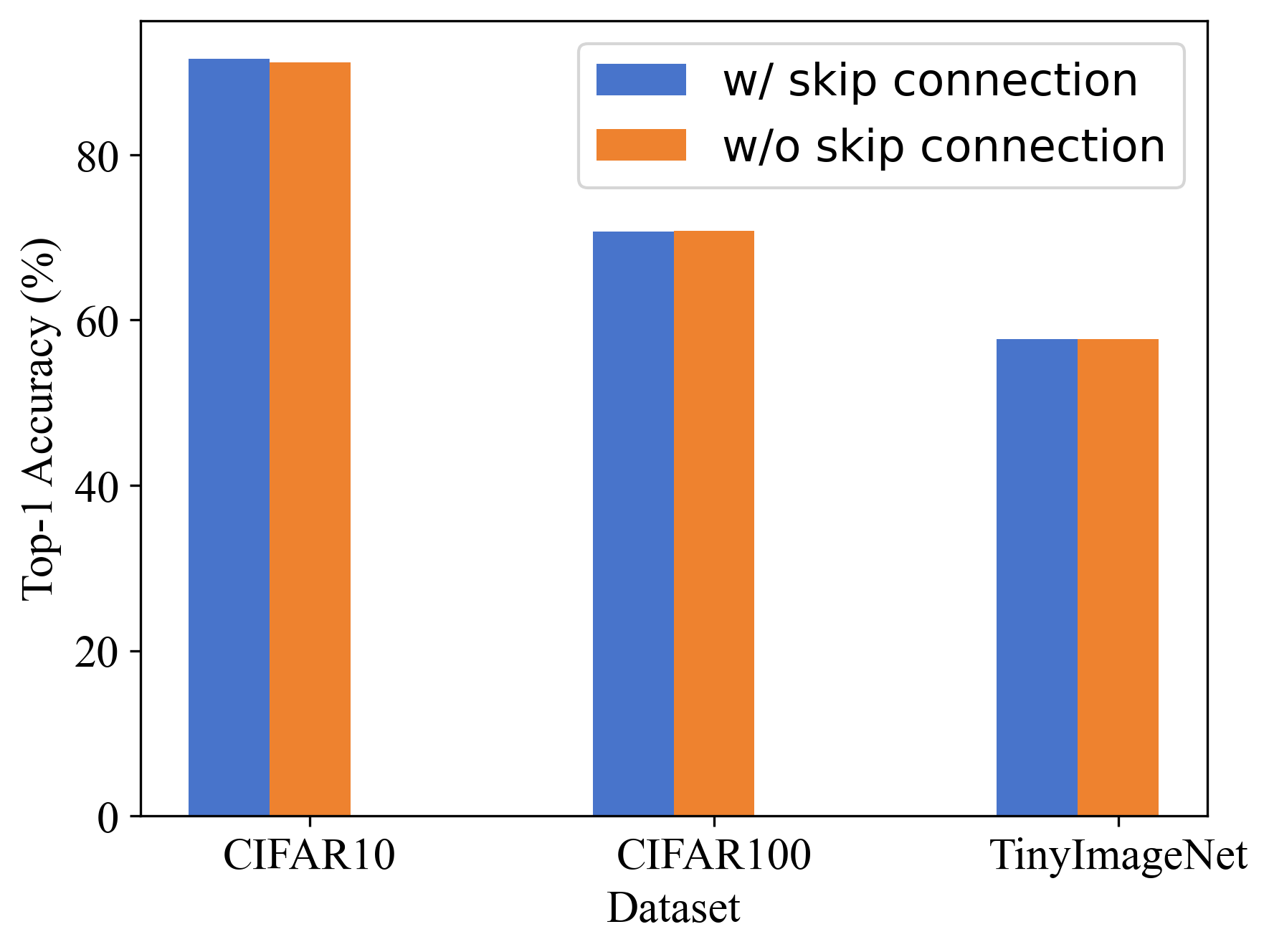}
    \caption{Top-1 Accuracy using the ConvMixer-Tiny model on
three different datasets with and without skip connections.}
    \label{fig:conv}
\end{figure}

In \cref{fig:conv}, we demonstrate that the performance of the ConvMixer Tiny model does not degrade when skip connections are removed. This observation contrasts with Vision Transformers (ViTs), where the absence of skip connections in the self-attention mechanism leads to a noticeable performance drop.



\section{Empirical Neural Tangent Kernel}
\begin{figure}[b]
    \centering
    \includegraphics[width=1\linewidth]{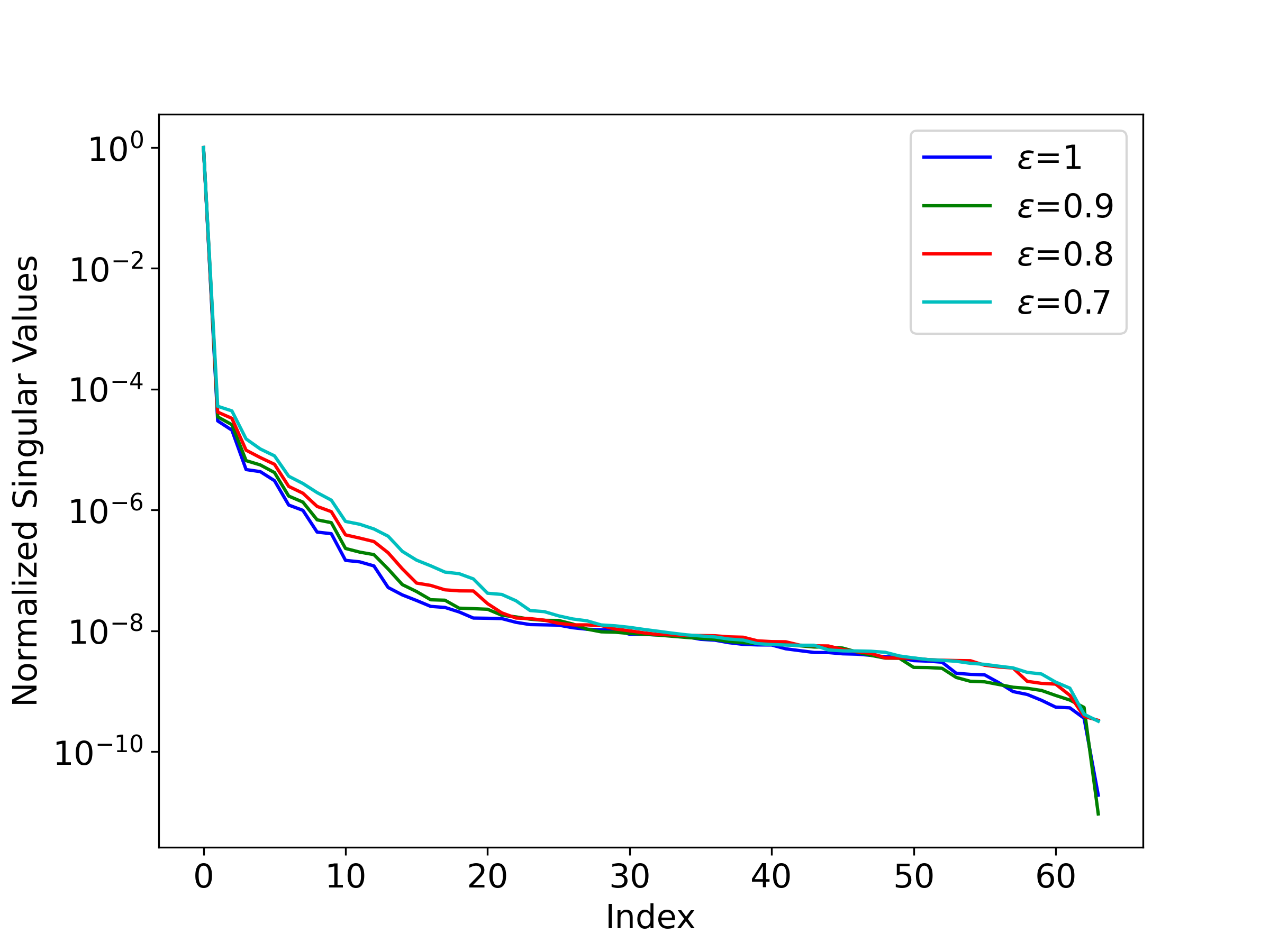}
    \caption{Spectrum of Jacobian of one SAB using \textbf{SVDTG}. Lower $\epsilon$ represents to better input token condition.}
    \label{fig:ntksvd}
\end{figure}

\begin{figure}[t!]
    \centering
    \includegraphics[width=1\linewidth]{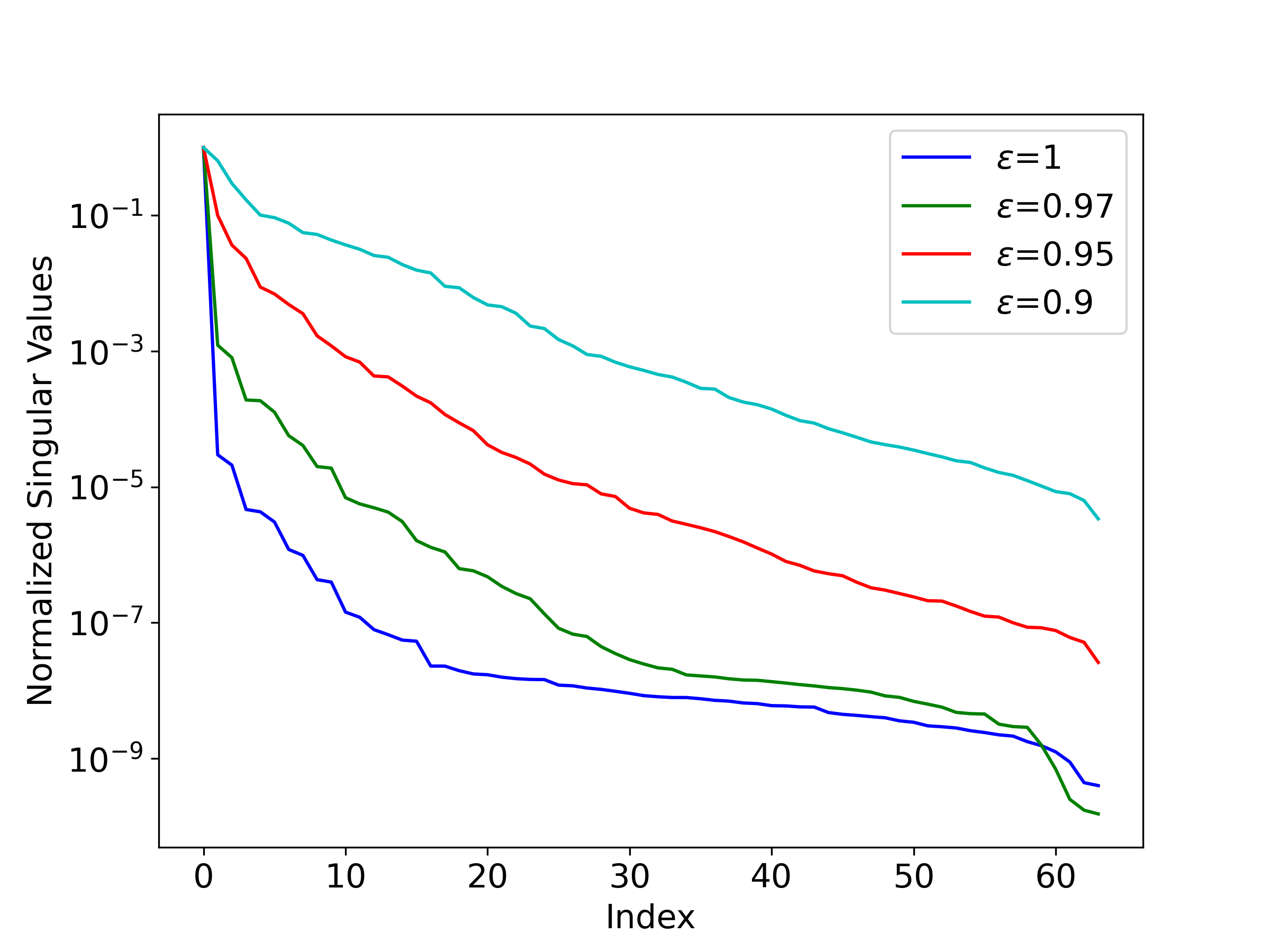}
    \caption{Spectrum of Jacobian of one SAB using \textbf{DCTTG}. Lower $\epsilon$ represents better input token condition.}
    \label{fig:ntkdct}
\end{figure}

The Neural Tangent Kernel (NTK) describes the evolution of deep neural networks during training by gradient descent. In this paper, we propose to measure the condition of the self-attention output embeddings as a proxy for its Jacobian condition, since analyzing the transformer model NTK is computationally expensive and unrealistic. However, in \cref{fig:ntksvd} and \cref{fig:ntkdct}, we empirically demonstrate the Jacobian of the self-attention mechanism using SVDTG and DCTTG. Intrinsically, the self-attention exhibits a disproportionately ill-conditioned spectrum, and using our TG methods, we observe an improvement in this regard.

\label{NTK}

\end{document}